\documentclass[lettersize,journal]{IEEEtran}
\usepackage{amsmath,amsfonts}
\usepackage{algorithmic}
\usepackage{algorithm}
\usepackage{array}
\usepackage[caption=false,font=normalsize,labelfont=sf,textfont=sf]{subfig}
\usepackage{textcomp}
\usepackage{stfloats}
\usepackage{url}
\usepackage{verbatim}
\usepackage{graphicx}
\usepackage{cite}
\begin{document}

\title{Being Aware of Localization Accuracy By Generating Predicted-IoU-Guided Quality Scores}

\author{Pengfei Liu, Weibo Wang, Yuhan Guo, Jiubin Tan
\thanks{This paper was produced by the IEEE Publication Technology Group. They are in Piscataway, NJ.}
\thanks{This work is supported by National Natural Science Foundation of China (52275527, 51975161 and 52275526); Key Research and Development Program of Heilongjiang (Grant No. 2022ZX01A27);CGN -HIT Advanced Nuclear and New Energy Research Institute (CGN-HIT202201).

Pengfei Liu, Weibo Wang and Jiubin Tan are with the Key Lab of Ultra-precision Intelligent Instrumentation (Harbin Institute of Technology), Ministry of Industry and Information Technology, Harbin 150001, China. (Corresponding author: WeiBo Wang, email:wwbhit@hit.edu.cn)

Yuhan Guo is with the School of Management, Harbin Institute of Technology, Harbin 150001, China}}
\markboth{IEEE TRANSACTIONS ON IMAGE PROCESSING,~Vol.~xx, No.~xx, JUNE~2023}%
{Shell \MakeLowercase{\textit{et al.}}: A Sample Article Using IEEEtran.cls for IEEE Journals}

\IEEEpubid{0000--0000/00\$00.00~\copyright~2021 IEEE}

\maketitle

\begin{abstract}
Localization Quality Estimation (LQE) helps to improve detection performance as it benefits post processing through jointly considering classification score and localization accuracy. In this perspective, for further leveraging the close relationship between localization accuracy and IoU (Intersection-Over-Union), and for depressing those inconsistent predictions, we designed an elegant LQE branch to acquire localization quality score guided by predicted IoU. Distinctly, for alleviating the inconsistency of classification score and localization quality during training and inference, under which some predictions with low classification scores but high LQE scores will impair the performance, instead of separately and independently setting, we embedded LQE branch into classification branch, producing a joint classification-localization-quality representation. Then a novel one-stage detector termed CLQ is proposed. Extensive experiments show that CLQ achieves state-of-the-arts’ performance at an accuracy of 47.8 AP and a speed of 11.5 fps with ResNeXt-101 as backbone on COCO test-dev. Finally, we extend CLQ to ATSS, producing a reliable 1.2 AP gain, showing our model's strong adaptability and scalability. Codes are released at \url{https://github.com/PanffeeReal/CLQ}.
\end{abstract}

\begin{IEEEkeywords}
Localization Quality Estimation, Object Detection, Soft Labels
\end{IEEEkeywords}

\section{Introduction}
\IEEEPARstart{A}{s} a fundamental while challenging task in computer vision, object detection aims to classify and localize objects-of-interest from normal images and it has many downstream applications. Since classification and regression branches are trained independently by two distinct loss functions, there occurs the inconsistency problem of the two \cite{ref1,ref2,ref3,ref4,ref5,ref6,ref7,ref8,ref9,ref10,ref11,ref12,ref13,ref14}. As some works indicate \cite{ref1,ref2,ref3,ref4,ref5}, selecting final predictions merely based on classification scores while ignoring the localization quality leads to a fact that some false positives predictions with high classification scores but low localization accuracies are selected during Non-Maximum Suppression (NMS) \cite{ref16}, which impairs the performance. Thus, Localization-Quality-Estimation (LQE) based or soft-weight-based methods \cite{ref1,ref4,ref8,ref13,ref17} are proposed to consist the two by introducing an independent IoU prediction branch or weighting the training classification loss \cite{ref17,ref18}, where the same insight is utilizing sample's IoU as localization quality \cite{ref2,ref4,ref19,ref20,ref21,ref22,ref23,ref24}. Following this direction, in this paper we design a lightweight LQE branch to densely predict feature points’ IoUs supervised by standard binary cross entropy (BCE) loss, making the detection head acquire the preliminary ability to depress those false positives.

Localization quality estimation takes ride of the close relationship with IoU \cite{ref4,ref11,ref15,ref23}, while their correlation is usually not linear \cite{ref4,ref6}. So simply taking predicted IoU as localization quality score is not precise. Then behind of IoU prediction, we further use an exponential function which takes predicted IoUs as inputs to produce the final IoU-guided localization quality scores, giving model’s ability to produce a suitable estimation score of prediction’s localization accuracy.

While as common practices do \cite{ref1,ref4,ref13,ref14,ref23}, LQE branch is usually parallel with classification or regression branch and the predicted localization quality score is multiplied by classification score during inference phase [1,4,13], which could only partly resolve the inconsistency problem of false positives because of the separate design and veiled solution \cite{ref2}, under which some predictions with low classification score but high localization quality scores would potentially rank in front of true positives and degrades the performance, which we call it the true negative problem. Thus instead of separately and implicitly using LQE branch, some other works \cite{ref2,ref3,ref6,ref24} tried to solve this problem through using a soft label or joint representation, whose insight is optimizing training process by replacing traditional one-hot classification targets with combining prediction’s classification score and its correlated IoU, which enables the classification branch to aware localization accuracy then forces the consistency of the two branches by directly using IoU as localization quality score. But this solution is rough and the process should be optimized. Distinctly, CLQ produces the joint classification-localization-quality score by merging the above LQE branch with classification branch and taking the product of the two predictions as final ranking scores used for NMS, where not only the false positive but also the true negatives could be suppressed by making consistent of classification, regression and localization quality branches. 

On the other hand, compared with two-stage detectors, one-stage counterparts maintain faster speed but lower accuracy partly because of directly doing dense prediction and lacking ROI proposals \cite{ref25}, which forces classification and regression to use misaligned features. Then various models are proposed\\ \hspace*{\fill} \\\cite{ref26,ref27,ref28,ref29,ref30} to adjust features before regression by introducing DCN \cite{ref29}. As a one-stage detector, there also needs a feature alignment branch before the joint classification-localization-quality branch, which could help further improve the performance. 

Finally, we built and trained our one-stage detector CLQ. Experiments show that CLQ achieves 47.8 AP at a speed of 11.5 fps with ResNeXt-101 as the backbone, surpassing most symbolic one-stage counterparts. We also extend CLQ to symbolic ATSS \cite{ref14}, producing a reliable 1.2 AP gain, which is a good improvement, showing CLQ’s strong adaptability and scalability.

\section{Related Works}
\subsection{One Stage Detectors}
Compared with two stage detectors \cite{ref31,ref32,ref33,ref34}, one-stage counterparts \cite{ref13,ref20,ref21,ref30,ref35} own faster inference speed but lower accuracy because of directly predicting classification confidence scores and localizations, where occurs a false positive  problem \cite{ref5,ref10,ref15,ref30,ref31}. To solve this problem, Consistent Optimization \cite{ref9} focuses on matching the training hypotheses and the inference quality by utilizing of the refined anchors during training. GA \cite{ref34} leverages semantic features and adapts features based on predicted anchors’ scales and aspect ratios at different locations. RefineDet \cite{ref26,ref27} propose a novel detection architecture, where the object detection module(ODM) further refines positive anchors predicted by anchor refine module(ARM) based on features aligned by transfer connection block(TCB). FoveaBox \cite{ref30} adopts DCN based on learned box offsets to improve its performance. AlignDet \cite{ref36} combines the flexibility of learned anchors and the preciseness of aligned features. As CLQ belongs to one-stage detector series, in this paper we also follow a design of feature alignment branch for CLQ based on predicted anchors and DCN to improve the performance.
\subsection{Localization Quality Estimation Based Detectors}
As classification confidence score is usually be treated as a reflection of predictions’ qualities \cite{ref1,ref2,ref3,ref6,ref13}, making some false positives [5,10] with high classification scores but low localization quality be selected during inference, which inevitably degrades the performance. To alleviate this problem, forked with some feature selection models \cite{ref5,ref10,ref13,ref30,ref37} aiming to select better positive samples across and within FPN feature levels, LQE-based methods aim to enable detection process to aware both classification and localization accuracy. IoU-Net \cite{ref1} and IoU-aware \cite{ref4} learn to predict the IoU between each detected bounding box and the matched ground-truth to acquire the score of localization accuracy, thus well localized bounding boxes are preserved in the NMS procedure. FCOS \cite{ref13} and ATSS \cite{ref14} proposed “Centerness” to lower samples’ contributions that are far away from center of positive regions, treating distance a reflection of localization quality. GFLV2 \cite{ref3} performs LQE using Distribution-Guided-Quality-Prediction module based on the learned distributions of the four parameters of the bounding box. Beyond the independent design of LQE branch that may also encounter an inconsistent of classification and localization quality \cite{ref2}, some other works modulate classification loss by adding samples’ IoUs as weight factors such as Consistent Loss \cite{ref8}, Noisy Anchors \cite{ref6}. More directly, some soft-label-based methods \cite{ref2,ref3,ref6,ref24} change the classification scores’ targets by a function of classification score and IoU during training. Inspired by all these excellent works, for well making consistent of classification and localization-quality, and taking a ride of the close relationship between LQE and IoU, we propose the CLQ detector where a novel joint classification-localization-quality representation is designed through combining classification and predicted-IoU-guided quality scores generated by proposed LQE branch.

\section{Method}
In this section, we introduce the proposed detector CLQ by comparing the novelty of our detection head with other symbolic LQE-based heads in section 3.1, detailed detection pipeline including training and inference in section 3.2, and the further improvement of CLQ with a feature alignment branch is in section 3.3. The novelties between our work and other related works are concluded as:
\begin{enumerate}{}{}
	\item{We designed a lightweight LQE branch to make consistent of classification score and localization quality by generating propriate IoU-guided LQE scores, based on which the detector is able to better aware the localization accuracy then selects well localized predicted bounding boxes without sacrificing inference speed.}
	\item{We merged the LQE branch with classification branch and produced an optimized joint ranking score, which could help further alleviate the inconsistency problem of classification and localization quality branches with better compressing inconsistent predictions during training and inference.}
	\item{We proposed a novel one-stage detector called CLQ, which achieves a 47.8 AP at a reference speed of 11.5 fps with ResNeXt-101 as backbone on COCO test-dev. And through extending our design to ATSS, a reliable 1.2 AP gain is acquired, showing our model’s broad application.}
\end{enumerate}
\subsection{Motivation and design of CLQ detection head}
As one key part, detection head is responsible for interested objects’ classification and localization, and followed by the one-stage detectors’ designing direction of dense prediction, false positives \cite{ref5,ref9,ref10} or the inconsistency of the two tasks caused by feature misalignment \cite{ref9,ref26,ref27,ref31,ref34,ref36} or the separate design of the two tasks \cite{ref1,ref2,ref13,ref14} becomes a key problem.
\begin{figure*}[!htb]
	\centering
	\includegraphics[width=6.5in]{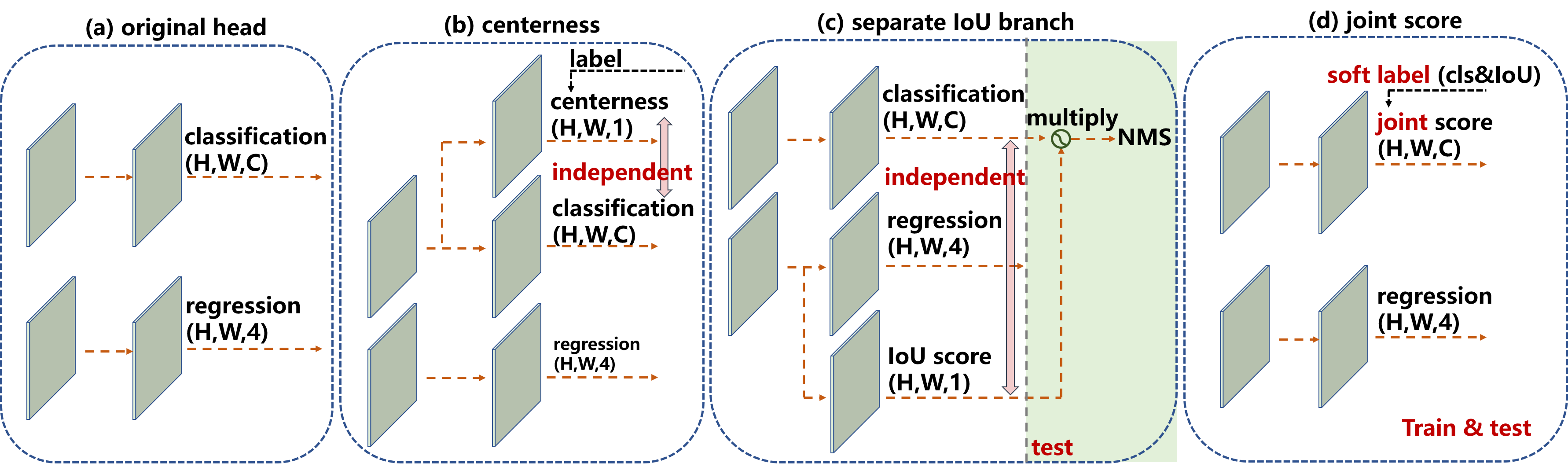}
	\caption{Previous symbolic detection heads.}
	\label{fig_1}
\end{figure*}

As Fig.1 shows, the original detection head (a) consists of two separate branches that have a weak correlation, leading to a fact that the predicted objects’ classification score and localization quality may diverge in training and inference phase, then some poor predictions with low localization accuracy or false positives may be selected as final results during post process \cite{ref16}. Then primary localization quality branch is proposed, as Fig.1(b) and (c) shows, the former utilize “centerness” score to lower poor samples’ contributions that are far away from the center of objects. The latter treats predicted IoU as localization quality, which is then combined with classification score to form a final score for the rank process of NMS. But this design brings another gap between training and inference, where some true negatives with a high localization quality score may rank in front of true positives then degrade model’s performance because of the separate and inconsistent usage of classification and localization quality branches \cite{ref2}. 

\begin{figure}[!htb]
	\centering
	\includegraphics[width=2.5in]{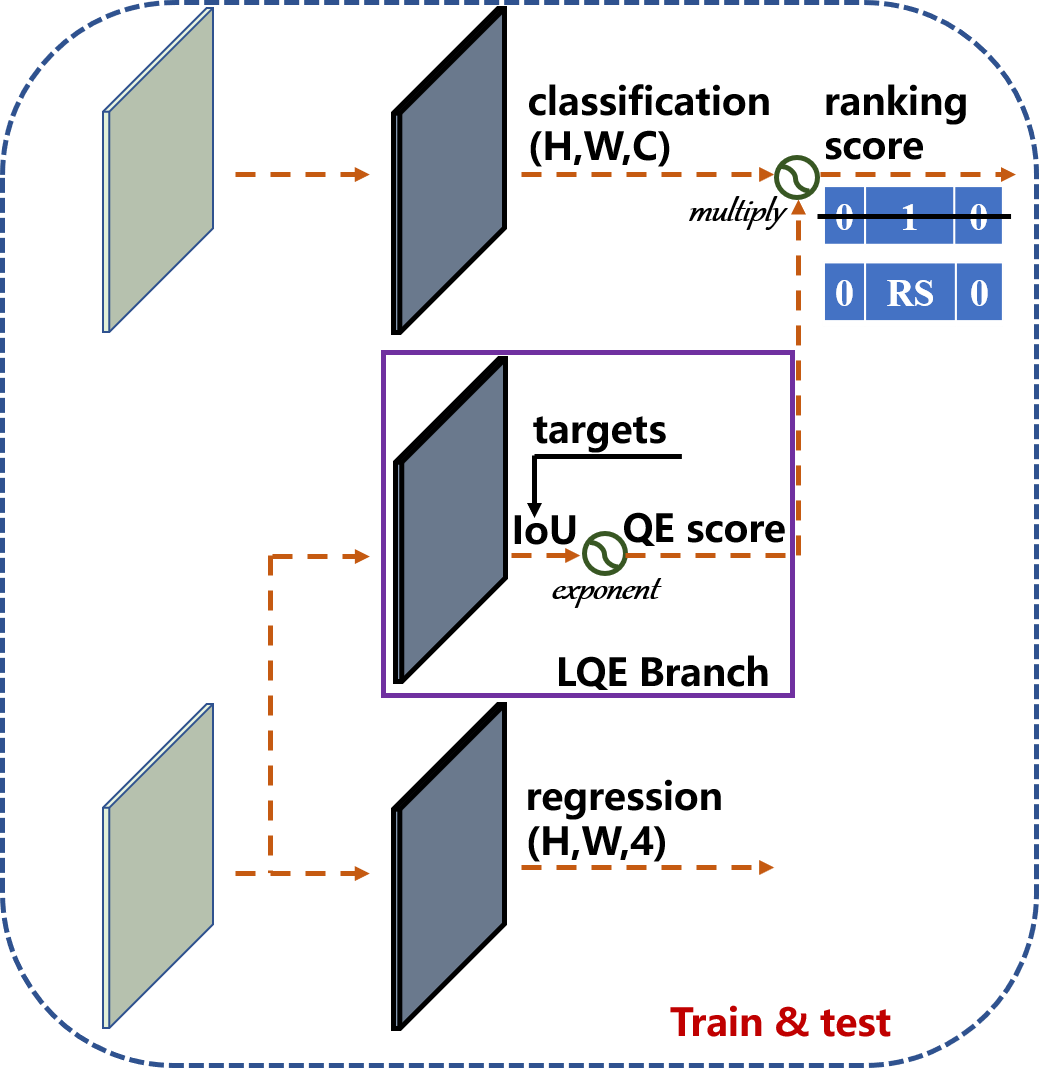}
	\caption{CLQ detection head.}
	\label{fig_2}
\end{figure}

For eliminating the above inconsistency risk while still being able to aware localization accuracy during inference, soft-label-based methods are proposed as Fig.1 (d) shows. Instead of using independent localization accuracy branch’s scores during inference, a joint score consisting of category score and samples’ IoUs are used to replace standard one-hot category label to train the classification branch, which realizes that being able to aware localization accuracy by regulating classification branch’ training. However, as this design directly takes IoU as soft labels, the classification branch will face more pressure and it could not show full strength because during inference the IoUs cease to exist. So we still believe the necessity of localization quality branch and there should exist a more reasonable instruction to generate better ranking scores.

Thus, we designed and proposed our CLQ head. As Fig.2 shows, distinct from previous soft-label-based LQE methods where QE score is constituted directly by classification score and samples’ IoUs during training, or some independent QE branch which only works during inference phase, we improved the detection head by firstly designing an IoU-guided LQE branch responsible for predicting a reliable QE score, then eliminating the gap caused by separate design and making it effective both during training and inference phase by merging the QE score with classification score. In the meantime, considering the efficiency, the LQE branch only bring a projection layer connected with regression branch, which brings a negligible computation burden to the whole model, and a sigmoid activation layer to ensure a reasonable predicted IoU value between 0 and 1. 

As for the training procedure, BCE loss is used to supervise the LQE branch and positive samples are used for calculating the loss, as (1) indicates.
\begin{equation}
\label{eq1}
	L_{lqe} = \frac{1}{N_{pos}}\sum_{1}^{N_{pos}}\text{BCEloss}(IoU_i,IoU_i'),
\end{equation}
where $IoU_{i}'$ is the training target defined by the overlap of current anchor box $b_{i}$ and its corresponding ground-truth box. Instead of directly using predicted IoU as QE score roughly during inference, we define the QE score using its exponential form as (2) indicates.
\begin{equation}
	\label{eq2}
	QE\_score_{i} = IoU_{i}^\alpha
\end{equation}

Secondly, for better making consistence of classification and localization quality, we adopt the joint representation to produce final ranking score (RS) to help NMS select truly good predictions, as (3) indicates.
\begin{equation}
ranking\ score_{i} = cls\_score_{i} *  QE\_score_{i},
\end{equation}
where the $cls\_score_{i} = [C_{1},C_{1},...C_{M}], C_{i}\in[0,1]$ stands for the classification predictions, and $QE\_score_{i}$  means the corresponding i\textsuperscript{th} localization quality score. The final ranking score is taken as the input of post process for the prediction results.
\subsection{Training and inference pipeline of CLQ}
The proposed detector follows one-stage detector’s design and the forward pipeline consists of a backbone to extract primary features, an FPN to build an enhanced feature pyramid, and the proposed detection head to realize dense prediction for classification and regression, producing classification results and 4-dimensional class-agnostic offsets for every feature point. The whole framework is shown in Fig.3.
\begin{figure*}[!htb]
	\centering
	\includegraphics[width=6.0in]{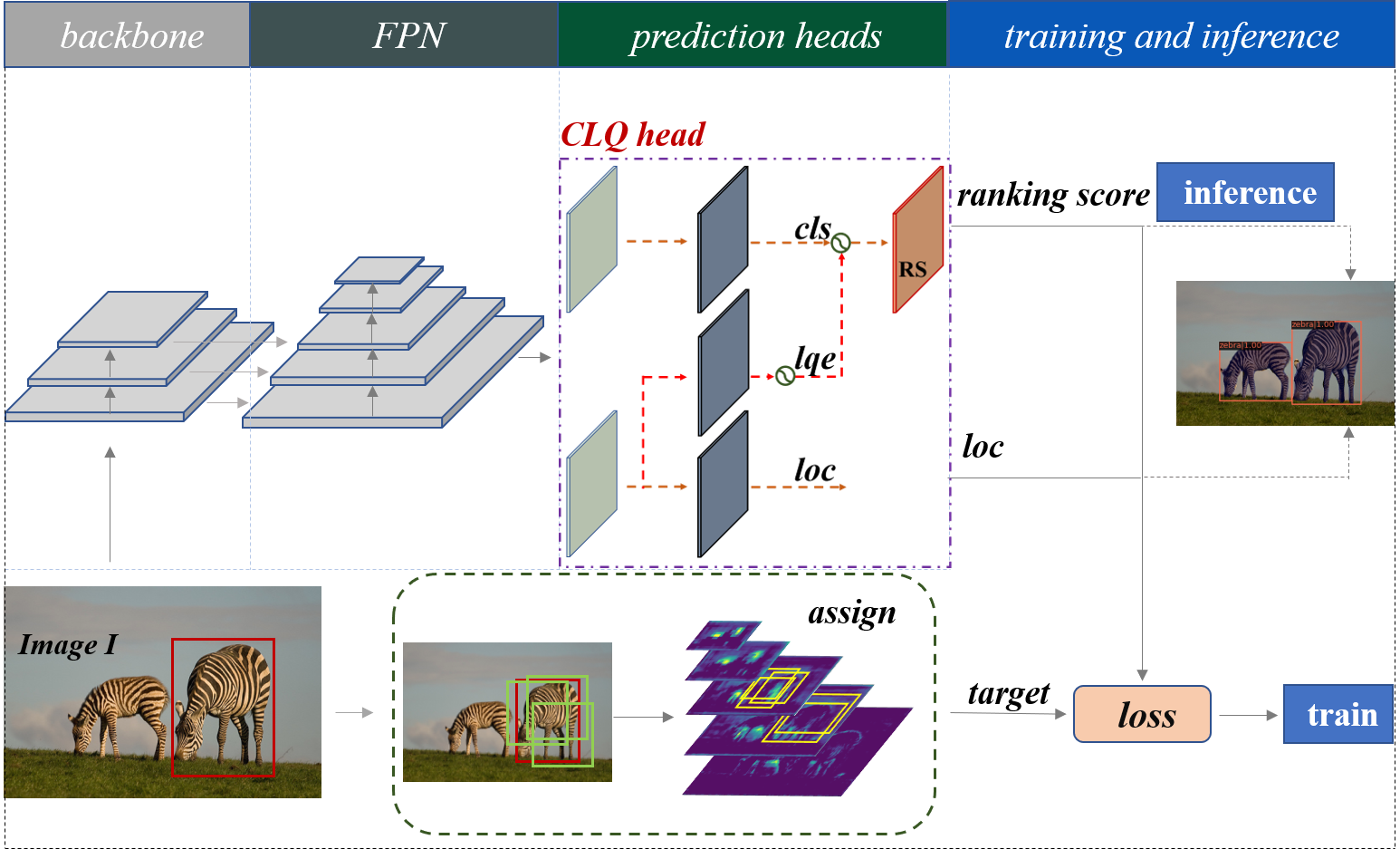}
	\caption{Training and inference pipeline of CLQ.}
	\label{fig_3}
\end{figure*}

Then given an image $I$, the assigner first selects positive anchors from the feature pyramid and generates corresponding targets for training, where standard ATSS \cite{ref14} assigner is adopted with selecting positive samples according to statistical characteristics of objects. Then the training loss is calculated by Quality-Focal-Loss \cite{ref2} and GIoU Loss \cite{ref22} respectively for classification and regression branches, as (4) and (5) indicates.

\begin{equation}
	L_{cls} = \frac{1}{N_{pos}}\sum_{1}^{N_{total}}\text{QFL}(ranking\ score_{i},(cls_{i}',QE\_target_{i})),
\end{equation}
\begin{equation}
	L_{reg} = \frac{1}{N_{pos}}\sum_{1}^{N_{pos}}\text{GIoU}(loc_{i},loc_{i}'),
\end{equation}
\begin{equation}
	L_{total} = L_{cls} + L_{reg} + L_{lqe},
\end{equation}
where $loc_{i}$ is predicted localization, $loc_{i}'$ is corresponding ground-truth box, $ranking\ score_{i}$ is the joint representation of classification and QE score, $cls_{i}'$ is category label, and $QE\_target_{i}$ is the training target of QE score. For negative feature sample, the corresponding category label and $QE\_target$ are set to be zero. Finally, with the aforementioned additional LQE branch’s loss, classification and regression loss, the total loss function is calculated as (6) indicates.

\subsection{Further improvement based on feature alignment}
As a one-stage anchor-based detector, CLQ directly predicts category scores based on features within preset anchor boxes, while compared to which the regressed bounding boxes have changed a lot, causing feature misalignment and increasing the number of false positives. As Fig.4 shows, original red anchor contains less foreground feature so classification branch tends to classify it as background. But according to the close relationship with the black ground-truth box, the regressed yellow box gets closer to the black box and should be classified as person, where the contradiction affects the performance.
\begin{figure}[!htb]
	\centering
	\includegraphics[width=3.0in]{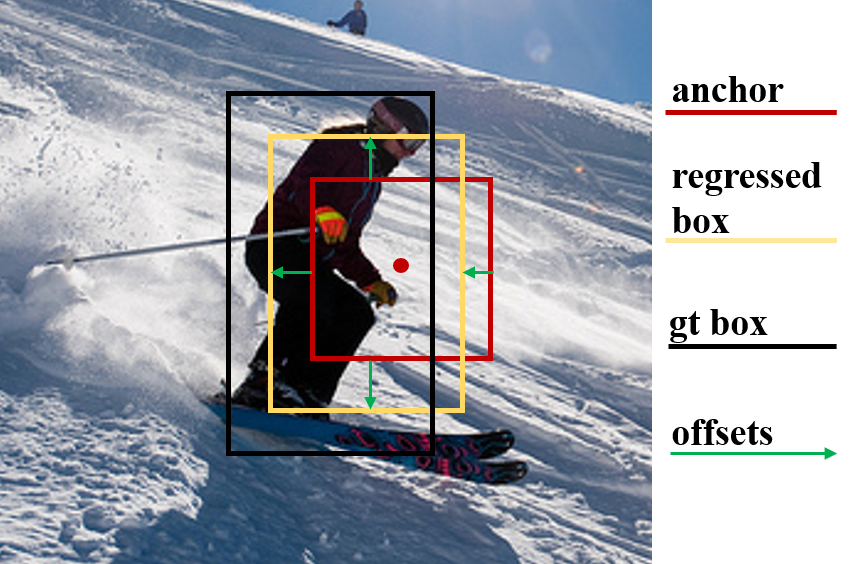}
	\caption{Description of feature misalignment.}
	\label{fig_4}
\end{figure}

To solve this problem, we further refined the CLQ head and adopted a DCN \cite{ref29} component for using aligned features based on regression branch’s offset predictions. As a result, the final ranking scores consisting of more accurate classification scores and LQE scores could better indicate the localization qualities of the regressed bounding boxes, thus improves model’s performance. The structure of improved CLQ head is shown in Fig.5.
\begin{figure}[!htb]
	\centering
	\includegraphics[width=3.0in]{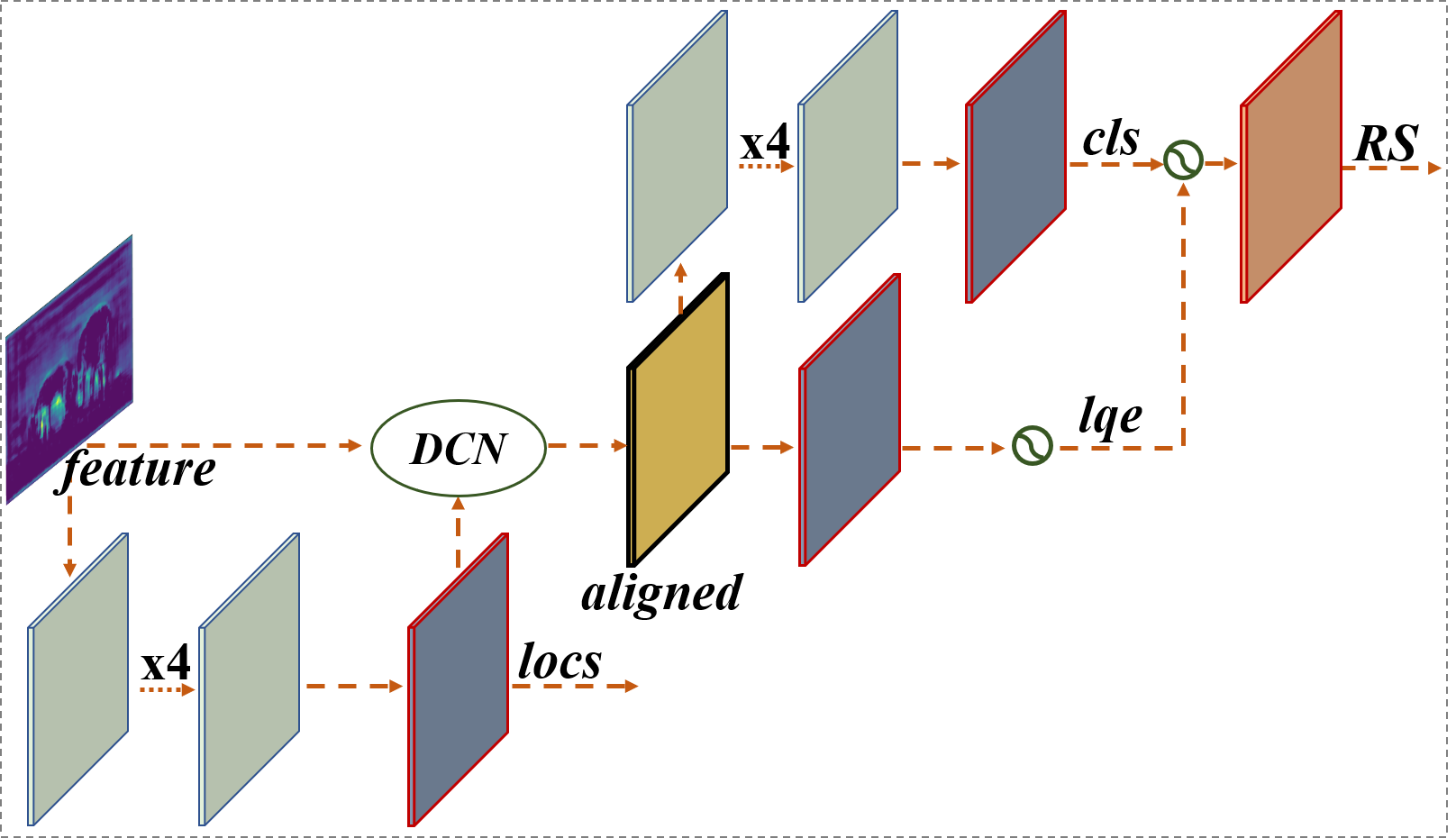}
	\caption{Design of improved CLQ head based on DCN.}
	\label{fig_5}
\end{figure}

Taking the original features from FPN and densely predicted regression offsets as inputs, DCN adjusts features under the guidance of regressed boxes’ localizations to ensure the feature regions used for classification and localization quality estimation well match the corresponding regressed bounding box.

\section{Experiments}
\subsection{Experimental Details}

Dataset settings: We did experiments on COCO benchmark \cite{ref38}, where trainval35k with 115K images are used for training and minival with 5K images for validation. To demonstrate the generalization ability of our detector, the standard COCO-style Average Precision (AP) metrics are adopted.

Implementation Details: All the object detectors are implemented based on Pytorch and MMdetection \cite{ref39} for fair comparison where the default hyper-parameters are adopted. Unless otherwise stated, we apply the standard 1x learning schedule (12 epochs) with a single shorter size of 800 pixels and without multi-scale training for the ablation study, based on ResNet-50 \cite{ref40} as backbone.
\begin{table}[!htb]
	\caption{The effect of different alpha\label{tab:table1}}
	\centering
	\begin{tabular}{c|c c c c c c }
		\hline 
		$\alpha$ & $AP$ & $AP_{50}$	& $AP_{75}$ & $AP_{S}$ & $AP_{M}$ & $AP_{L}$\\
		\hline
		0.2 & 40.2 & 58.2 &	43.5 & 22.5 & 44.0 & 52.3\\
		0.3 & 40.4 & 58.3 &	43.7 & 23.4 & 44.4 & 52.8\\
		0.4 & 40.2 & 58.2 &	43.5 & 23.2 & 44.1 & 52.6\\
		\hline
	\end{tabular}
\end{table}
\begin{table*}[!b]
	\caption{The effect of distinct branches\label{tab:table2}}
	\centering
	\begin{tabular}{c|c c c | c c c c }
		\hline 
		model   & qfl         & lqe	&  align    & fps   & AP    & $AP_{50}$ & $AP_{75}$\\
		\hline
		CLQ      & \checkmark  &        &           & 24.7 & 39.9 & 58.5 & 43.0\\
		\text{w lqe}  & \checkmark &\checkmark& & 24.5 & $40.4_{+0.5}$ & 58.3 & 43.7\\
		w lqe \& align  & \checkmark &\checkmark&\checkmark & 23.7 & $41.6_{+1.7}$ & 59.5 & 45.3\\
		\hline
	\end{tabular}
\end{table*}
\subsection{Ablation Study}
Accurate localization quality score generation: we first explore our model with various $\alpha$ during training and the results are shown in Table.1. For generating more suitable localization quality scores, we use the hyper-parameter $\alpha$ to modulate the contributions of LQE branch. As (2) shows, where bigger $\alpha$ makes the quality score play a bigger role while vice versa. Based on Table.1, we choose $\alpha=0.3$  as optimized value, which provides a higher 0.2 AP improvement, indicating an unequal contribution of classification and LQE branch when generating the final ranking scores. 

CLQ head improves detection performance: to respectively verify the effect of the designed branches in CLQ head, we further did ablation study by adding corresponding modules one by one, and results are shown in Table.2. As CLQ takes a joint representation of category score and localization quality score as final ranking score, where the standard one-hot category label is softened as continuous values, QFL \cite{ref2} which is suitable for continuous category loss’s calculation is used to replace traditional Focal Loss \cite{ref35}. Based on that, as Fig.2 shows, adding an LQE branch into vanilla one stage detection head in our CLQ model gives a 0.5 AP improvement with almost no sacrifice in speed because of the negligible amount of extra parameters. When further using aligned features, a total 1.7 AP improvement can be acquired at the expense of one seconds inference time, which is 23.7 fps, but this inference speed still meets most real time requirements

\begin{table}[!h]
	\caption{Extension of CLQ to ATSS\label{tab:table3}}
	\centering
	\begin{tabular}{c|c c c | c c c }
		\hline 
		model   & qfl         & lqe	&  align     & AP    & $AP_{50}$ & $AP_{75}$\\
		\hline
		\text{ATSS}   &    & & & 39.4 & 57.6 & 42.8 \\
		
		\text{ w qfl}      & \checkmark  &        &     & $39.8_{+0.4}$ & 57.5 & 43.5\\
		\text{ w CLQ}  & \checkmark &\checkmark&  & $40.2_{+0.8}$ & 58.1 & 43.8\\
		\text{ w CLQ}  & \checkmark &\checkmark&\checkmark  & $40.6_{+1.2}$ & 58.5 & 44.5\\
		\hline
	\end{tabular}
\end{table}
Good scalability of CLQ: beyond CLQ, we transfer the designed LQE branch and feature alignment module to symbolic one stage detector ATSS. Based on the results shown in Table.3, adding an LQE branch into ATSS gives a 0.4 AP improvement compared with only using QFL, and when further using aligned features, an extra 0.8 AP improvement can be reached. Totally, a 1.2 AP improvement can be provided with the help of CLQ head. More detailed, as shown in Fig. 6, the training loss curves with different epochs are drawn. We conclude the CLQ does improve detection performance, indicating its good effectiveness and scalability.

\begin{figure}[!htb]
	\centering
	\includegraphics[width=3.5in]{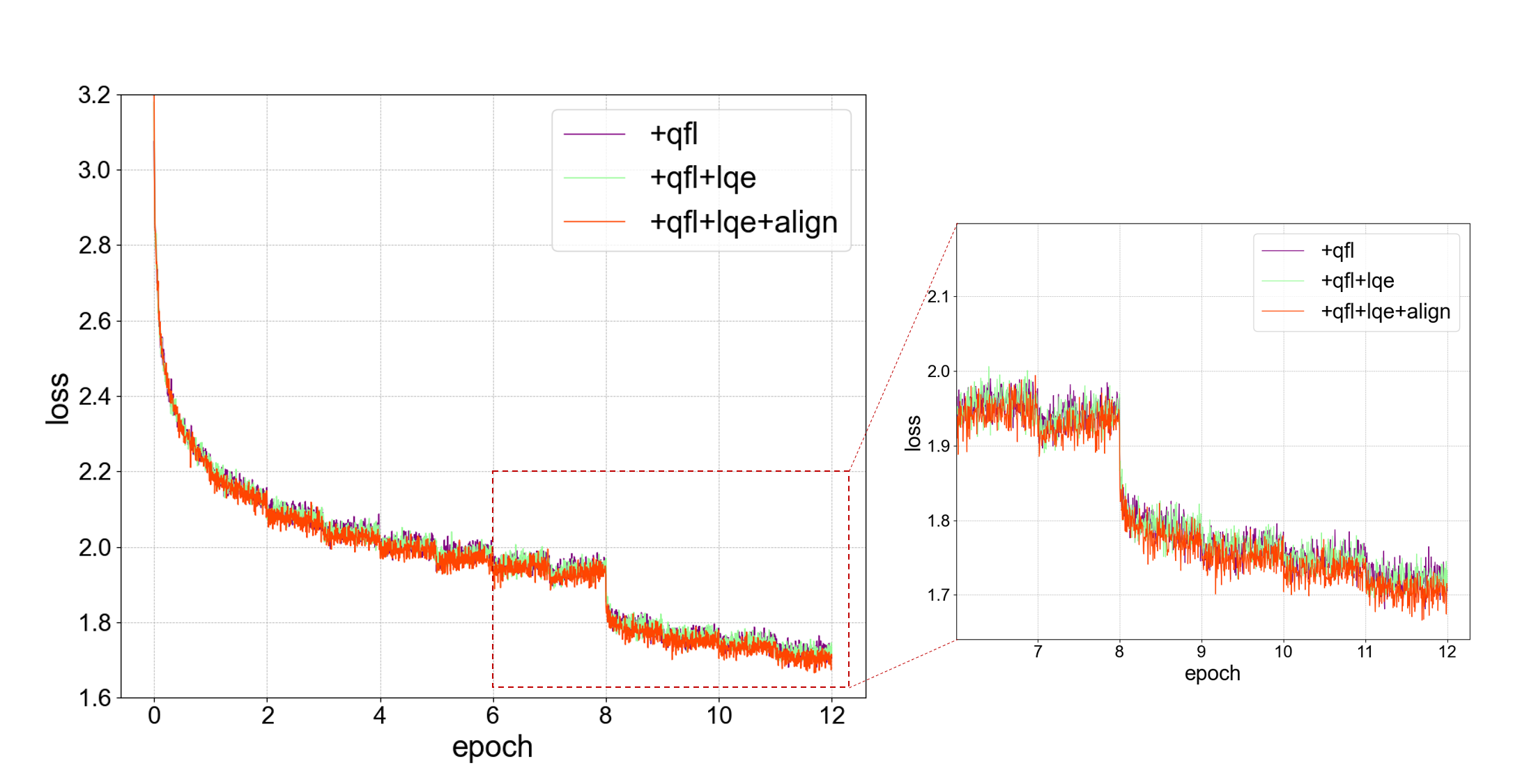}
	\caption{Ablation study of training loss with ATSS as the base model.}
	\label{fig_6}
\end{figure}

\begin{table*}[!ht]
	\caption{comparison with the state-of-the-arts on COCO test-dev. FPS values with “*” are from papers \cite{ref3,ref5}. while others are measured on the same machine with a single GeForce RTX 3090 GPU and Intel Xeon E5-2630L v3 CPU. “R”: ResNet\cite{ref40}. “X”: ResNeXt\cite{ref41}. “HG”: Hourglass\cite{ref42}.\label{tab:table4}}
	\centering
	\begin{tabular}{c|c| c c c c c c| c |c |c }
		\hline 
		model &backbone&AP&$AP_{50}$&$AP_{75}$&$AP_{S}$&$AP_{M}$&$AP_{L}$&fps&epoch&Anchor Free?\\
		
		\hline
		\text{CornerNet}\cite{ref43} & HG-104& 40.5&56.5&57.8&19.4&	42.7&53.9&$3.1^*$&200&y \\
		\text{Centernet}\cite{ref44} & HG-104& 42.1&61.1&45.9&24.1& 45.5&52.8&$3.3^*$&190&y \\
		\text{ExtremeNet}\cite{ref45}& HG-104& 40.2&55.5&43.2&20.4&	43.2&53.1&$2.8^*$&200&y \\
		\hline
		\text{Faster R-CNN}\cite{ref25}&R-101& 39.8&60.1&43.3&22.5&	43.6&52.8&	15.6&	24&	n \\
		\text{Faster R-CNN}\cite{ref25}&X-101-32x4d	&41.2&61.7&44.8&23.9&44.9&54.3&14.6&24&n \\
		\text{Mask R-CNN}\cite{ref32}&R-101&40.8&61.0&44.5&23.0&45.0&54.1&11.7&24&n \\
		\text{Mask R-CNN}\cite{ref32}&X-101-32x4d&42.2&62.6&46.0&23.5&46.1&56.0&11.4&24&n \\
		\text{GAFasterRCNN}\cite{ref34}&R-101&41.5&60.5&45.3&23.3&45.4&55.6&12.1&12&n \\
		\text{GAFasterRCNN}\cite{ref34}&X-101-32x4d&42.9&62.5&46.9&24.6&47.4&56.1&11.5&12&n \\
		\text{Cascade R-CNN}\cite{ref31}&	R-101&42.5&60.7&46.4&23.5&46.5&56.4&13.5&20&n \\
		\text{Cascade R-CNN}\cite{ref31}&	X-101-32x4d&44.1&62.6&48.0&24.8&47.9&57.2&12.8&20&n \\
		\hline
		\text{FCOS}\cite{ref13}&	R-101&41.5&60.2&43.6&24.2&44.1&52.1&$14.7^*$&24&y \\
		\text{FCOS}\cite{ref13}&	X-101-64x4d&44.7&62.5&45.7&26.0&46.5&54.7&$8.9^*$&24&y \\
		\text{Foveabox}\cite{ref30}&	R-101&40.8&61.4&44.0&24.1&45.3&53.2&16.1&24&y \\
		\text{SAPD}\cite{ref5}&	R-50&41.7&61.9&44.6&24.1&44.6&51.6&$14.9^*$&24&y \\
		\text{SAPD}\cite{ref5}&	R-101&43.5&63.6&46.5&24.9&46.8&54.6&$13.2^*$&24&y \\
		\text{SAPD}\cite{ref5}&	X-101-32x4d-DCN&46.6&66.6&50.0&27.3&49.7&60.7&$8.8^*$&24&y \\
		\text{LRD}\cite{ref37}&	R-101&42.7&62.2&46.4&25.4&47.2&54.7&15.5&24&y \\
		\text{LRD}\cite{ref37}&	X-101-64x4d&44.9&64.7&48.4&27.3&48.6&58.1&11.6&24&y \\
		\text{FSAF}\cite{ref10}&	R-101&40.9&61.5&44.0&24.0&44.2&51.3&15.3&24&y \\
		\text{FSAF}\cite{ref10}&X-101-64x4d&42.9&63.8&46.3&26.6&46.2&52.7&11.3&24&y \\
		\text{AutoAssign}\cite{ref46}&	R-50&40.4&59.6&43.7&22.7&44.1&52.9&10.2&12&y \\
		\hline
		\text{Retinanet}\cite{ref35}&	R-101&38.9&58.0&41.5&21.0&42.8&52.4&14.4&24&n \\
		\text{Retinanet}\cite{ref35}&	X-101-32x4d&40.1&59.7&42.7&21.8&44.3&53.3&13.2&24&n \\
		\text{ATSS}\cite{ref14}&	R-101&43.6&62.1&47.4&26.1&47.0&53.6&$14.6^*$&24&n \\
		\text{ATSS}\cite{ref14}&	X-101-32x8d-DCN&47.7&66.6&52.1&29.3&50.8&59.7&$6.9^*$&24&n \\
		\text{FreeAnchor}\cite{ref47}&	R-101&43.1&62.2&46.4&24.5&46.1&54.8&13.0&12&n \\
		\text{FreeAnchor}\cite{ref47}&	X-101-32x4d&44.9&64.3&48.5&26.8&48.3&55.9&12.8&12&n \\
		\text{GFLV1}\cite{ref2}&	R50&43.1&62.0&46.8&26.0&46.7&52.3&18.4&24&n \\
		\text{GFLV1}\cite{ref2}&	R-101&44.9&63.1&49.0&28.0&49.1&57.2&15.5&24&n \\
		\text{GFLV1}\cite{ref2}&	X-101-32x4d&46.1&64.8&50.3&28.9&50.4&58.4&14.4&24&n \\
		\text{GFLV2}\cite{ref3}&	R-50&44.3&62.3&	48.5&26.8&47.7&54.1&18.8&24&n \\
		\text{GFLV2}\cite{ref3}&	R-101&46.2&	64.3&50.5&27.8&49.9&57.0&14.6&24&n \\
		\text{TOOD}\cite{ref48}&	R-101&46.7&	64.6&50.7&28.9&49.6&57.0&13.9&24&n \\
		\text{TOOD}\cite{ref48}&	X-101-64x4d&48.3&66.5&52.4&30.7&51.3&58.6&10.2&24&n \\
		\text{VFNET}\cite{ref24}&	R50&44.3&62.5&48.1&26.7&47.3&54.3&16.8&24&n \\
		\text{VFNET}\cite{ref24}&	R-101&46.0&64.2&50.0&27.5&49.4&56.9&13.7&24&n \\
		\text{VFNET}\cite{ref24}&	X-101-32x4d&46.7&65.2&50.8&28.3&50.1&57.3&13.3&24&n \\
		\hline
		\text{CLQ}&	R-50&44.2&63.1&48.1&25.7&47.4&54.7&18.3&24&n \\
		\text{CLQ}&	R-101&46.2&64.6&50.6&27.7&49.9&57.8&15.1&24&n \\
		\text{CLQ}&	X-101-64x4d&47.8&66.4&52.0&29.3&51.0&60.1&11.5&24&n \\	
		\hline
	\end{tabular}
\end{table*}
\subsection{Comparison with the state of the arts}
We evaluate our model on the COCO test-dev to compare with symbolic state-of-the-art convolution-based object detectors. Multi-scale training strategy and 2x learning schedule (24 epochs) are adopted. For fair comparison, the trained models 
\begin{figure}[!htb]
	\centering
	\includegraphics[width=3.0in]{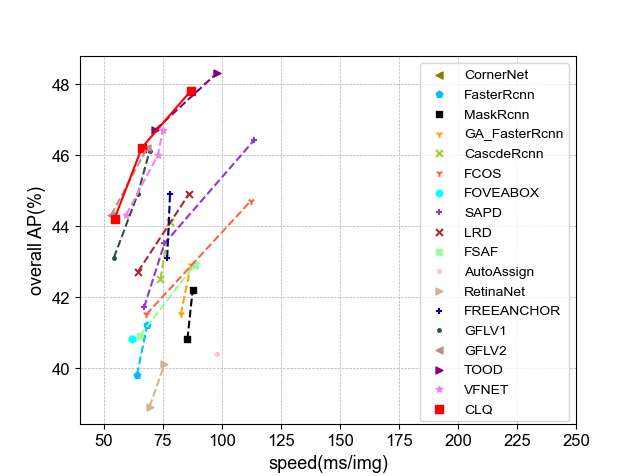}
	\caption{Speed(ms) vs accuracy(AP) on COCO test-dev. We compare CLQ(our model) with other symbolic detectors. CLQ reaches the state-of-the-arts’ performance with a (AP, speed) of (44.2, 18.3), (46.2,15.1), (47.8,11.5) based on ResNet-50, ResNet-101, ResNeXt-101 respectively.}
	\label{fig_7}
\end{figure}
provided by MMDetectioin \cite{ref39} with different backbones are evaluated as the baselines. The comparison results are shown in Table 4, our model performs well and it reaches the state-of-the-arts’ performance, and our best result with a single ResNeXt-101-64x4d model achieves considerably competitive 47.8 AP with an inference speed of 11.5 fps.

For better visualization of the speed and accuracy trade off, we graphically represent the data in Table 4, as Fig.7 shows. Because of the elegant design with extra negligible parameters, CLQ offers a significant speed advantage while maintains competitive accuracy, and produces a (AP, speed) of (44.2, 18.3), (46.2,15.1), (47.8,11.5) based on ResNet-50, ResNet-101, ResNeXt-101 respectively, showing a broader application potential by guarantying the real-time performance.

As a concluding, the designed CLQ generates a suitable localization quality score and works together with the classification branch, making the detector able to consider localization quality both during training and inference, thus improving the consistency of original independent classification and localization tasks. Moreover, the performance is further improved by using aligned features taking the ride on the coattails of deformable convolutional network. Fig.8 shows detection results of CLQ based on ResNet-101, with the designed CLQ, an accurate classification and localization ability can be reached, especially for small or densely distributed objects.
\subsection{Discussions}
\begin{figure*}[!hbt]
	\centering
	\includegraphics[width=6.5in]{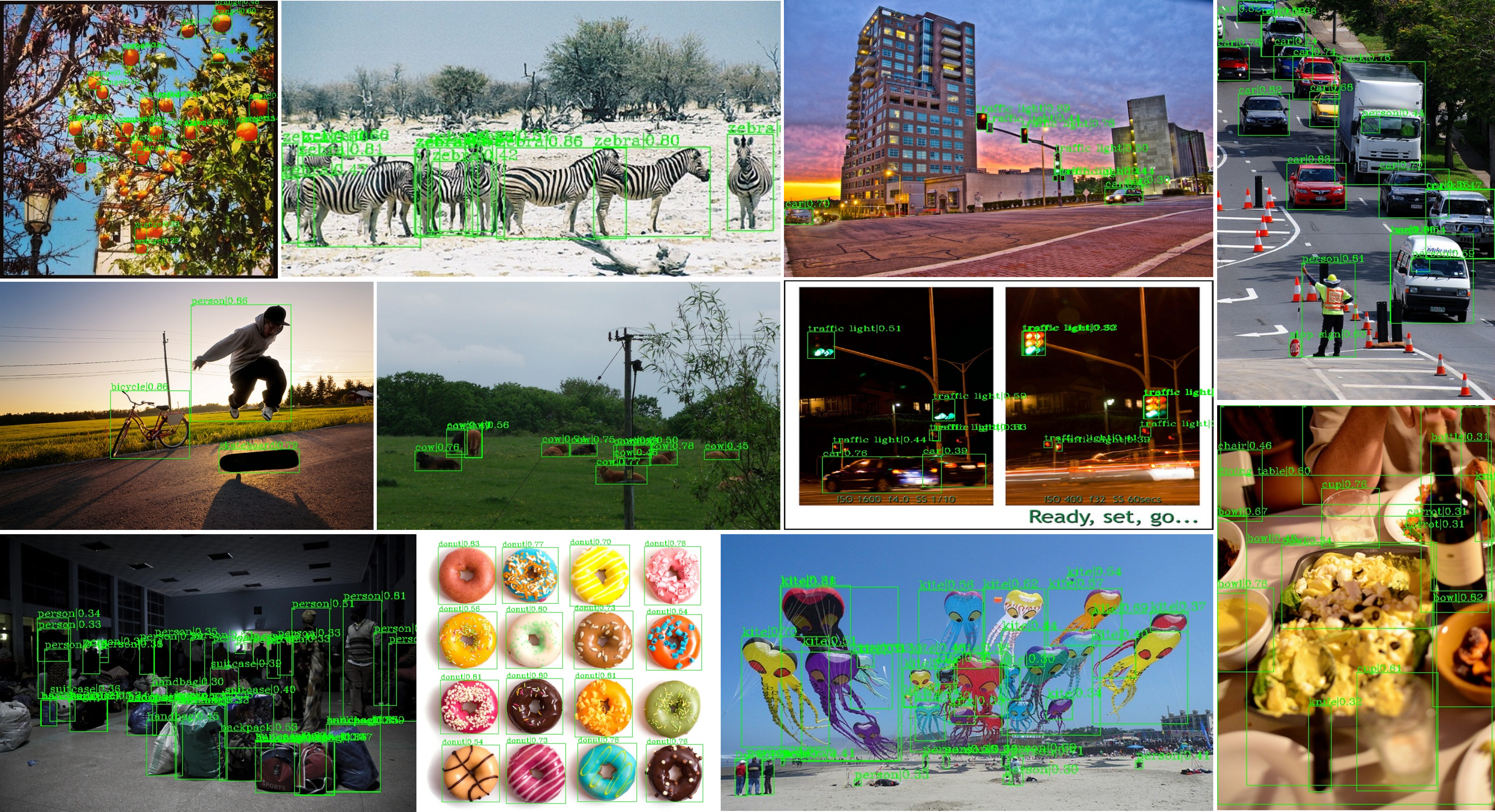}
	\caption{Visualization of CLQ’s inference results based on ResNet-101. Our model shows good ability on both classification and localization.}
	\label{fig_8}
\end{figure*}

Localization quality estimation is an effective mean to help NMS select better predictions with more accurately localized bounding boxes by merging its quality score with original classification scores. While the quality score has a close relationship with IoU which relies heavily on anchors. On the other side, anchor-free counterparts usually maintain faster speed and competitive detection accuracy because of less anchor-related parameters \cite{ref13} and region-based label assignment \cite{ref5,ref10,ref30,ref37,ref46}. Currently, among anchor-free detectors, there is few localization quality estimation related works. Thus, improving anchor-free detectors by riding on the coattails of the localization quality estimation to further raise the envelope of accuracy-speed boundary may become another potential research. Finally, as improving the detector without sacrificing inference speed is always our goal, CLQ is proposed with an elegant LQE branch and a lightweight feature alignment branch, and we plan to convert it to anchor-free detectors to further boost the inference speed.
\section{Conclusions}
In this paper we propose a new one-stage detector called CLQ, which generates proper localization quality scores guided by predicted IoU with the help of designed lightweight LQE branch without sacrificing inference speed, and optimizes the joint representation by embedding this LQE branch with classification branch instead of directly taking positive samples’ IoUs with ground-truth boxes as quality scores. Then the detector is able to well aware predicted boxes’ localization accuracy during both training and inference. Moreover, we add a lightweight feature alignment branch guided by predicted regression offsets based on DCN to further improve CLQ’s performance. Finally, based on the ablation experiment results on COCO dataset and through extending CLQ to symbolic ATSS detector, the efficiency and compatibility are validated, showing our model’s potential on broader applications.

\end{document}